\documentclass[letterpaper, 10 pt, conference]{ieeeconf}  

\IEEEoverridecommandlockouts                              

\usepackage{graphicx}
\usepackage{amsmath}
\usepackage{amsfonts}
\usepackage{bm}
\usepackage{booktabs}
\usepackage{color}
\usepackage{url}
\usepackage{amsmath}
\usepackage{algorithm}
\usepackage{algpseudocode}
\usepackage{mdframed}
\usepackage{hyperref}
\usepackage{multirow}
\usepackage{comment}
\usepackage{makecell}

\title{\LARGE \bf
Modality-Driven Design for Multi-Step Dexterous Manipulation: Insights from Neuroscience
}
\author{Naoki Wake$^{1}$, Atsushi Kanehira$^{1}$, Daichi Saito$^{2}$, Jun Takamatsu$^{1}$, \\ Kazuhiro Sasabuchi$^{1}$, Hideki Koike$^{2}$, and Katsushi Ikeuchi$^{1}$%
\thanks{$^{1}$Applied Robotics Research,                Microsoft, Redmond, WA, USA}%
\thanks{$^{2}$Department of Computer Science,
        Institute of Science Tokyo, Tokyo, Japan}
}

\begin{document}
\maketitle
\thispagestyle{empty}
\pagestyle{empty}

\begin{abstract}
Multi-step dexterous manipulation is a fundamental skill in household scenarios, yet remains an underexplored area in robotics. This paper proposes a modular approach, where each step of the manipulation process is addressed with dedicated policies based on effective modality input, rather than relying on a single end-to-end model. To demonstrate this, a dexterous robotic hand performs a manipulation task involving picking up and rotating a box. Guided by insights from neuroscience, the task is decomposed into three sub-skills—reaching, grasping and lifting, and in-hand rotation—based on the dominant sensory modalities employed in the human brain. Each sub-skill is addressed using distinct methods from a practical perspective: a classical controller, a Vision-Language-Action model, and a reinforcement learning policy with force feedback, respectively. We tested the pipeline on a real robot to demonstrate the feasibility of our approach. The key contribution of this study lies in presenting a neuroscience-inspired, modality-driven methodology for multi-step dexterous manipulation.
\end{abstract}

\section{Introduction}
Multi-step dexterous manipulation plays a crucial role in many household tasks, yet it remains challenging for robots. For instance, a sequential task of grasping an object and changing its posture is a common action observed in daily life. However, even such a seemingly simple task involves several sub-tasks from a robotics perspective, including finding the object, approaching it, grasping, lifting, and manipulating in hand. Additionally, these sub-tasks must be sequenced with precise control and timing, making multi-step dexterous manipulation a significant challenge in the robotics community.

A recent prominent approach seeks to solve multi-step tasks using a single model. Specifically, ``Vision-Language-Action (VLA) models'' (e.g., ~\cite{kim2024openvla,brohan2023rt,brohan2022rt}) generate step-wise robotic motions based on visual and verbal information. However, deploying the VLA models in the real world presents several issues. Possible task branches emerge as the sequence length increases, making it extremely challenging to gather training data covering all possible scenarios. Additionally, current VLA models are trained with RGB images and teleoperation data using parallel-jaw grippers (e.g.,~\cite{o2023open}), thus limiting their application to tasks that involve additional modalities and action complexity. For example, visual input often involves self-occlusions during operations, necessitating the use of other modalities such as tactile or force feedback. In other cases, some grasps cannot be achieved with conventional parallel-jaw grippers and need dexterous manipulators that come with increased action dimensions.

\begin{figure}[t]
  \centering
  \includegraphics[width=\columnwidth]{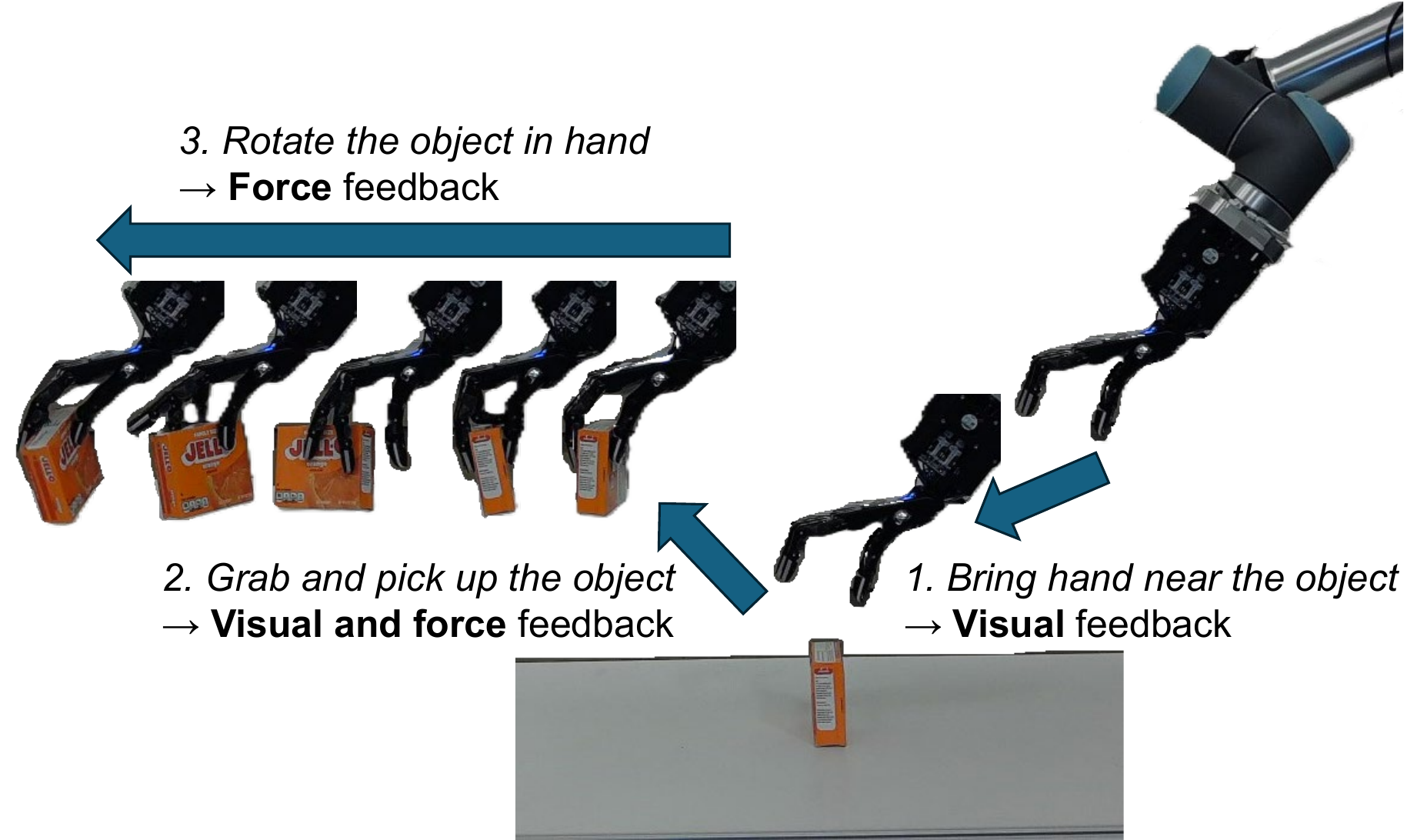}
  \caption{In this paper, we propose that multi-step dexterous manipulation can be addressed by sequencing sub-skills tailored to specific sensory modalities. We focus on a manipulation task that involves picking up and rotating a box. Guided by neuroscience evidence, this task is decomposed into three sub-tasks: approaching the object using visual feedback, grasping and lifting using visual and force feedback, and rotating the object in hand using force feedback.}
  \label{fig:topfigure}
\end{figure}

Toward scalable and efficient multi-step dexterous manipulation, we advocate decomposing tasks into sub-tasks and addressing them with suitable control methods, rather than relying on a single model (Fig.~\ref{fig:topfigure}). This decomposition approach is not new; recent solutions in robotic manipulation have proposed leveraging Large-Language-Models (LLMs) and Vision-Language-Models (VLMs) to understand and decompose human instruction into detailed sub-steps (e.g., ~\cite{xu2023creative,zhou2023generalizable,sun2024prompt,wake2024gpt,wake2023chatgpt}), in the context of task and motion planning~\cite{garrett2021integrated}. Cheng et al.~\cite{pan2024vision} and Mehta et al.~\cite{mehta2024feasibility} have demonstrated that a multi-step approach achieves higher success rates in dexterous manipulation compared to solving with a single model. However, the choice of appropriate modalities and learning approaches for dexterous manipulation has not been sufficiently investigated.

This paper aims to provide practical insights into the decomposition of multi-step tasks and the corresponding control methods for the resulting sub-tasks. We focus on the task of grasping and rotating a box with a dexterous robotic hand. Drawing on neuroscience perspectives, we decompose this task into reaching, grasping, and in-hand manipulation, based on the effective modalities required for each sub-task. Each sub-task is implemented using different methods, including classical controllers, visual feedback controllers, and RL policies with force feedback. 
The contributions of this paper are: 
\begin{itemize}
    \item Proposing a modality-driven design informed by neuroscience insights.
    \item Demonstrating multi-step dexterous manipulation through the sequencing of sub-skills in a real-robot setup.
    \item Providing practical guidelines for successful implementation, which include a vision-based teleoperation system and the use of augmented simulation data to enhance model robustness.
\end{itemize}

\section{Modality-Driven Design for Multi-Step Dexterous Manipulation}\label{method}
Although the action of grasping and manipulating an object may appear seamless in humans, behavioral experiments and brain imaging studies have revealed that different types of information are utilized in each phase. These insights provide valuable clues, particularly for robots with dexterity comparable to humans, regarding what type of information contributes to efficient motion in each phase. Therefore, we propose that considering the modalities involved in each phase is effective for devising task decomposition and determining appropriate learning methods for each sub-task.
\subsection{Reaching}
We define human reaching as the act of moving the hand close to an object without pre-shaping it, as preparation for a subsequent grasping task. This task relies on vision-based spatial awareness to approach the object without interacting with or colliding with the environment. Neuroimaging studies have shown that reaching engages cortical regions responsible for spatial recognition (e.g., superior parietal lobue~\cite{kawashima1995functional,grafton1996,connolly2000comparison,simon2002topographical}), spatial motion planning (e.g., dorsal premotor cortex~\cite{astafiev2003functional}),  and coding spatial locations (e.g., frontal eye fields~\cite{astafiev2003functional}), which reflect eye-position signals. \textbf{These findings emphasize the importance of spatial awareness and planning in reaching}. In robotics, reaching can thus be framed as a vision-based motion planning problem. While recent methods leverage VLMs to achieve spatial awareness~\cite{chen2024spatialvlm, huang2023voxposer}, classical controllers remain a robust alternative, effectively utilizing the spatial information of the workspace.

\subsection{Grasping and lifting}
We define grasping as the operation to achieve appropriate contact positions and the finger postures required to reach those positions. From both robotic and anatomical perspectives, grasps are categorized into precision and power types~\cite{napier1956prehensile, yoshikawa1999passive}. Precision grasping involves pinching an object between the fingers and the opposing thumb, while power grasping involves securely holding an object between the flexed fingers and the palm. One factor determining the grasp type is the task's purpose~\cite{napier1956prehensile}. Here, verbal input is utilized to understand the goal and intent.

Another factor influencing the grasp type is the object's shape and size~\cite{napier1956prehensile}, for which visual information is essential~\cite{saito2022task}, while verbal information may convey the object's affordance~\cite{wake2020grasp,wake2023text}. In the field of neuroscience, visually-guided grasping involves cortical areas that code object information (e.g., ventral premotor cortex~\cite{prabhu2009modulation, murata1997object}), suggesting that visual information plays critical role in grasping objects~\cite{sakata1995neural,borra2008cortical}. 

Another important aspect in grasping is adjusting the positional relationship between the object and the hand. Visual information is again suggested to be utilized to this end (e.g., dorsal premotor cortex~\cite{goodale1992separate,cisek2003neural}). Furthermore, the grasping task involves cortical regions where tactile sensations are integrated with visual information (e.g., superior parietal lobule~\cite{coull1996fronto,wolpert1998maintaining,grafton1996functional}), suggesting that proprioceptive information with the object is used in grasping behavior. In the context of the lifting task, studies have shown that humans adjust their grip force based on the object's surface properties and weight~\cite{johansson1984roles,flanagan2000independence}. Therefore, proprioceptive information, which conveys tactile and/or contact-force feedback from the environment, plays a critical role in lifting, especially when the object's weight and surface friction coefficient are unknown.

\textbf{Therefore, visual, verbal, and proprioceptive information are effective in object grasping and lifting.} Robot-control methodologies that leverage language and visual information can be broadly categorized into top-down and bottom-up approaches. The top-down approach includes Learning-from-Observation (LfO)~\cite{ikeuchi2024semantic,wake2021learning}. In this method, the grasp type is inferred from human multimodal demonstrations.  The finger configurations (i.e., contact web~\cite{saito2022task}) on object's surface are then determined visually using techniques such as CNNs. In contrast, bottom-up approaches, such as VLA models, allow direct action outputs from language instructions and visual inputs~\cite{kim2024openvla,team2024octo}.

These two approaches have pros and cons. The method of directly estimating the contact web has limitations when occlusion occurs during approach. In contrast, bottom-up methods may accommodate partial occlusions if those appear in the training data, but do not guarantee the coherent action outputs due to the non-unimodal nature in the latent space~\cite{wu2024discrete}. In this paper, we assume a hybrid method that a VLA model generates low-level grasping motions for a specific grasp type, following LfO that determines a grasp type from human demonstration (\cite{wake2024gpt}).

\subsection{In-hand manipulation}
\textbf{In object manipulation tasks that require fast cyclic movements and small finger displacements, tactile and force feedback are critical, whereas visual feedback is less important}~\cite{augurelle2003importance,nowak2003selective,monzee2003effects}. Visual feedback may not be utilized in real time, and visual occlusion can become an issue. From a neuroscience perspective, in-hand tasks involve the recruitment of somatosensory and motor areas compared to visual areas~\cite{binkofski1999fronto}. This observation aligns with findings from other in-hand manipulation studies in robotics, which demonstrate that tasks can be successfully performed without relying on visual information~\cite{van2015learning,pitz2024learning}. It is noteworthy that visual feedback may be effective depending on the complexity of the tool usage, such as tongs manipulation~\cite{inoue2001activation}. In this study, however, we consider in-hand rotations as contact-rich tasks where proprioceptive information is dominant and propose an RL approach that relies solely on force feedback.

\section{Related Work}
\subsection{Learning in-hand dexterous manipulation}
Research on in-hand dexterous manipulation has made significant progress~\cite{andrychowicz2020learning, handa2023dextreme, yin2023rotating, chen2022system, qi2023hand, qi2023general,arunachalam2023dexterous,akkaya2019solving,chen2022system}. Despite the widespread use of RL in this domain~\cite{chen2022system, andrychowicz2020learning,chen2022towards, kumar2016optimal,van2015learning}, its application to complex tasks remains challenging due to difficulties in reward design and long training times. To address this, recent studies decompose in-hand manipulation tasks to reduce learning complexity~\cite{li2024interactive, yang2024task, gupta2021reset}. Our approach follows this strategy by segmenting object rotation into finger movement patterns on the object's surface~\cite{saito2024apricot}, learning each segment with RL, and sequencing them, enabling efficient learning and reduced complexity. 
\subsection{Multi-step robot operation}
While studies on in-hand dexterous manipulation report successful results, most focus on single-step skills, assuming the object is already in hand. In contrast, we extend these methods by sequencing multiple skills, such as grasping and lifting. Skill sequencing offers the flexibility to adopt the most suitable modalities and learning methods for each skill, as discussed in Sec.~\ref{method}. Although prior studies have highlighted the benefits of this approach~\cite{pan2024vision, mehta2024feasibility, chen2023sequential}, we aim to further explore effective modalities for learning each skill informed by neuroscience studies.

\subsection{Teleoperation for dexterous manipulation}
Even if VLA models are pre-trained on large-scale data, they are still sensitive to environmental changes~\cite{wang2024towards} and limited to parallel-jaw grippers without dexterity. Therefore, model fine-tuning is necessary using dexterous manipulation demonstrations in specific environments. A common method of data collection is teleoperation. Teleoperation methods are broadly classified into glove-based~\cite{wang2024dexcap,schwarz2021nimbro,liu2017glove,liu2019high} and vision-based~\cite{sivakumar2022robotic,handa2020dexpilot,li2022dexterous,qin2022one,ding2024bunny,pan2024vision,qin2023anyteleop} approaches. Glove-based methods enable real-time capture of fine-grained movements and, in some cases, tactile feedback for the operator. In contrast, vision-based methods are more cost-effective and require minimal setup. Prior studies have demonstrated that vision-based approaches can effectively facilitate the teleoperation of dexterous robotic hands, for fundamental tasks such as object pickup~\cite{handa2020dexpilot}. Based on these considerations, we opted for a vision-based setup to simplify system deployment.

\section{Experiment}\label{experiment}
As shown by Cheng et al.~\cite{pan2024vision} and Mehta et al.~\cite{mehta2024feasibility}, a multi-step approach achieves higher success rates in dexterous manipulation compared to solving with a single model. Thus, we focus on the performance of our method rather than comparing with single-model approaches.
\subsection{Hardware setup}

We used a UR10e robot arm (Universal Robots)\footnote{https://www.universal-robots.com/products/ur10-robot/} with six degrees of freedom (DoF). A four-fingered robot hand, the Shadow Dexterous Hand Lite (Shadow Robotics)\footnote{https://www.shadowrobot.com/dexterous-hand-series/}, was attached to the robot. The robotic hand was not equipped with tactile sensors; instead, torque sensor values from one joint of each finger (FF3, MF3, RF3, and TH2, as defined in its official specifications\footnote{https://www.shadowrobot.com/dexterous-hand-series/}) were utilized as proprioceptive feedback for the system.

For visual feedback to the system, we used a ZED2i camera\footnote{https://www.stereolabs.com/products/zed-2} fixed in a position relative to the UR10e robot. Although the ZED2i is a stereo camera capable of calculating depth information, we used  only the left-side image because the VLA model used in this study~\cite{team2024octo} was pre-trained with RGB images.
\subsection{Data collection using teleoperation and fine-tuning}
In our system, a ZED2i stereo camera was used to calculate the 3D position of the wrist using its official SDK. Concurrently, a cropped image of the hand region was extracted and processed using an off-the-shelf hand reconstruction model~\cite{pavlakos2024reconstructing} to estimate the hand pose. By leveraging two NVIDIA RTX4090 GPUs, we achieved real-time tracking of wrist and finger positions and orientations at an approximate rate of 15 FPS.

\begin{figure}[t]
  \centering
  \includegraphics[width=\columnwidth]{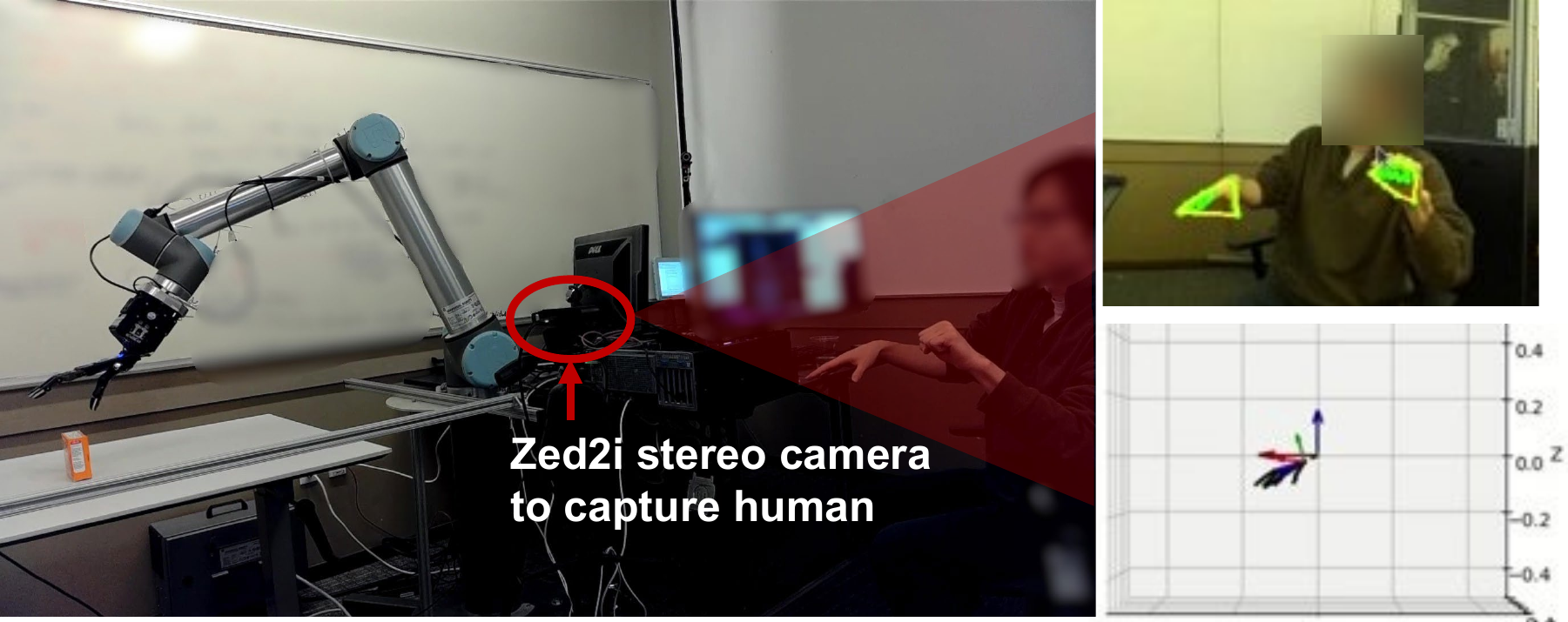}
  \caption{To collect human demonstrations, we developed a vision-based teleoperation system. In this system, right-hand motions were captured to control the robot arm and the Shadow hand, while left-hand gestures were used to start and stop the recording.}
  \label{fig:teleoperation}
\end{figure}

For data collection, the motion of the operator's right hand was mapped to the movements of the Shadow hand (Fig.~\ref{fig:teleoperation}). The initial wrist position was set using a gesture of the left hand, and subsequent relative left-wrist movements were reflected in the position of the Shadow hand. The Shadow hand's pose was fixed for precision grasping from above the object. The normalized distance between the operator's thumb and middle fingertip was linearly mapped to the closing degree of the robotic thumb and opposing fingers. The initiation and termination of data collection in each episode were signaled through predefined gestures of the left hand. 40 real-world demonstrations were collected with a gelatin box from the YCB object set~\cite{calli2015ycb}. 

\begin{figure}[t]
  \centering
  \includegraphics[width=\columnwidth]{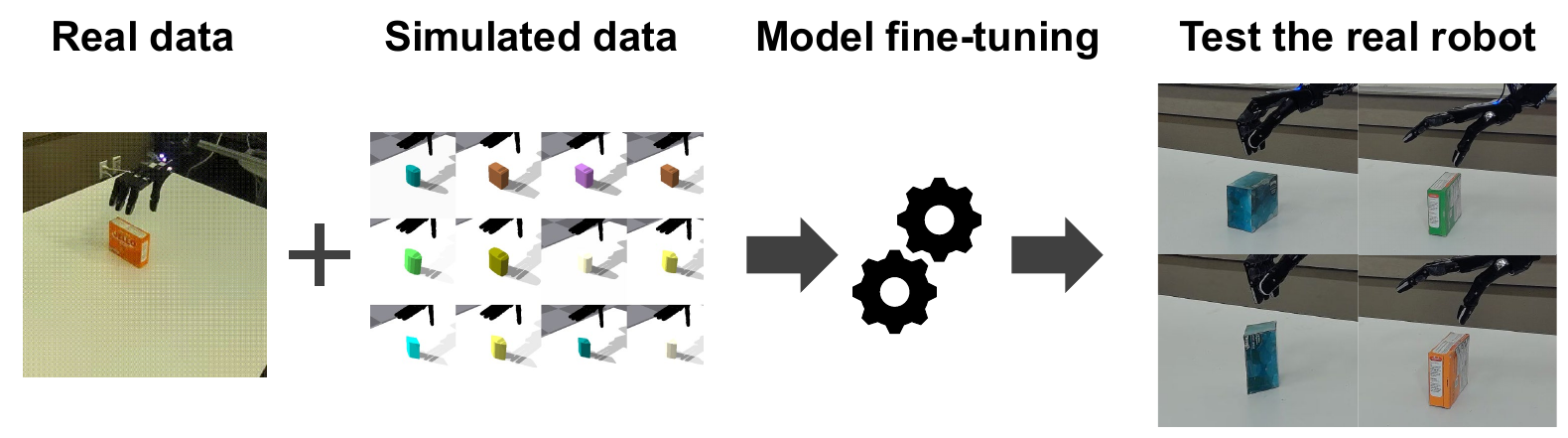}
  \caption{We prepared a mixed dataset consisting of real and simulated demonstrations to enhance the model's robustness.}
  \label{fig:finetuning}
\end{figure}

VLA models still face challenges in achieving robustness for practical robotics applications. Even in environments resembling their training demonstrations, factors such as lighting changes can lead to failures~\cite{wang2024towards}. While the diversity of environments and objects appears to play an important role in improving generalization performance~\cite{lin2024data}, capturing such diversity through real-world demonstrations is both time-consuming and costly. To address this, we adopt an approach that augments real-world datasets with simulated data. Specifically, we utilized IsaacGym to create a simulation environment replicating real-world conditions, introducing randomization in object position, shape, scale, and lighting (Fig.~\ref{fig:finetuning}). A pre-configured robot controller guided the robot hand and arm to contact points on the objects, based on precise 3D object positions within the simulation. This process yielded 4,000 simulated demonstrations, which were combined with the 40 real-world demonstrations.



\subsection{Classical control of robot hands}
We implemented inverse kinematics (IK) to determine the position and orientation of the robot hand. To accomplish the reaching task, the system first obtained the object's 2D position from a third-party object detector\footnote{https://azure.microsoft.com/en-us/products/ai-services/ai-custom-vision}, and then computed its 3D position using depth information from stereo cameras. This 3D position was transformed into the robot's coordinate system via an AR marker. The reaching target was defined as a predetermined relative position with respect to the object. During this process, the hand's orientation was maintained at a predefined pose. The IK solution was also applied at each action step determined by the Octo model.

\subsection{VLA model}
The current foundation models for robots primarily cannot be directly applied to our setup in off-the-shelf manner because the existing models and dataset focus on operating parallel jaw grippers. To adapt these models for dexterous manipulation, we utilized a model called Octo~\cite{team2024octo}. Octo features a mechanism known as block-wise attention structure, which allows for flexible modifications of the action space and observation space during fine-tuning. Instead of using the one-dimensional action space of a parallel gripper, we assigned multi-dimensional finger joint information. Based on the Virtual Finger theory~\cite{iberall1987grasp}, we imposed a constraint to synchronize three fingers opposing the thumb, reducing the 16-dimensional control parameters of the dexterous hand to 7 dimensions (4 dimensions for the thumb and 3 dimensions for the opposing fingers). Finally, we defined a 14-dimensional action space by incorporating 6 dimensions for controlling the hand's position and orientation, as well as 1 dimension to determine the termination of an episode. 

For the observation space, we extended the RGB image input by adding 4-dimensional torque information obtained from each finger. The captured RGB images were preprocessed before being input to the model by cropping the image to focus on the object. Specifically, the third-party object detector localized the object in the scene at the beginning of each episode, and subsequent images were cropped to a 400-pixel square region around the detected object, ensuring that both the object and the Shadow hand were visible in the frame (see Fig.~\ref{fig:finetuning} for an example image). This preprocessing was intended to create a focus-of-attention that removed irrelevant surrounding information while emphasizing the interaction between the object and the hand~\cite{wake2020verbal}. We included not just the immediately preceding frame but also the history of the past two steps to enrich the temporal context.

\subsection{RL for in-hand rotation}
\begin{figure}[t]
  \centering
  \includegraphics[width=\columnwidth]{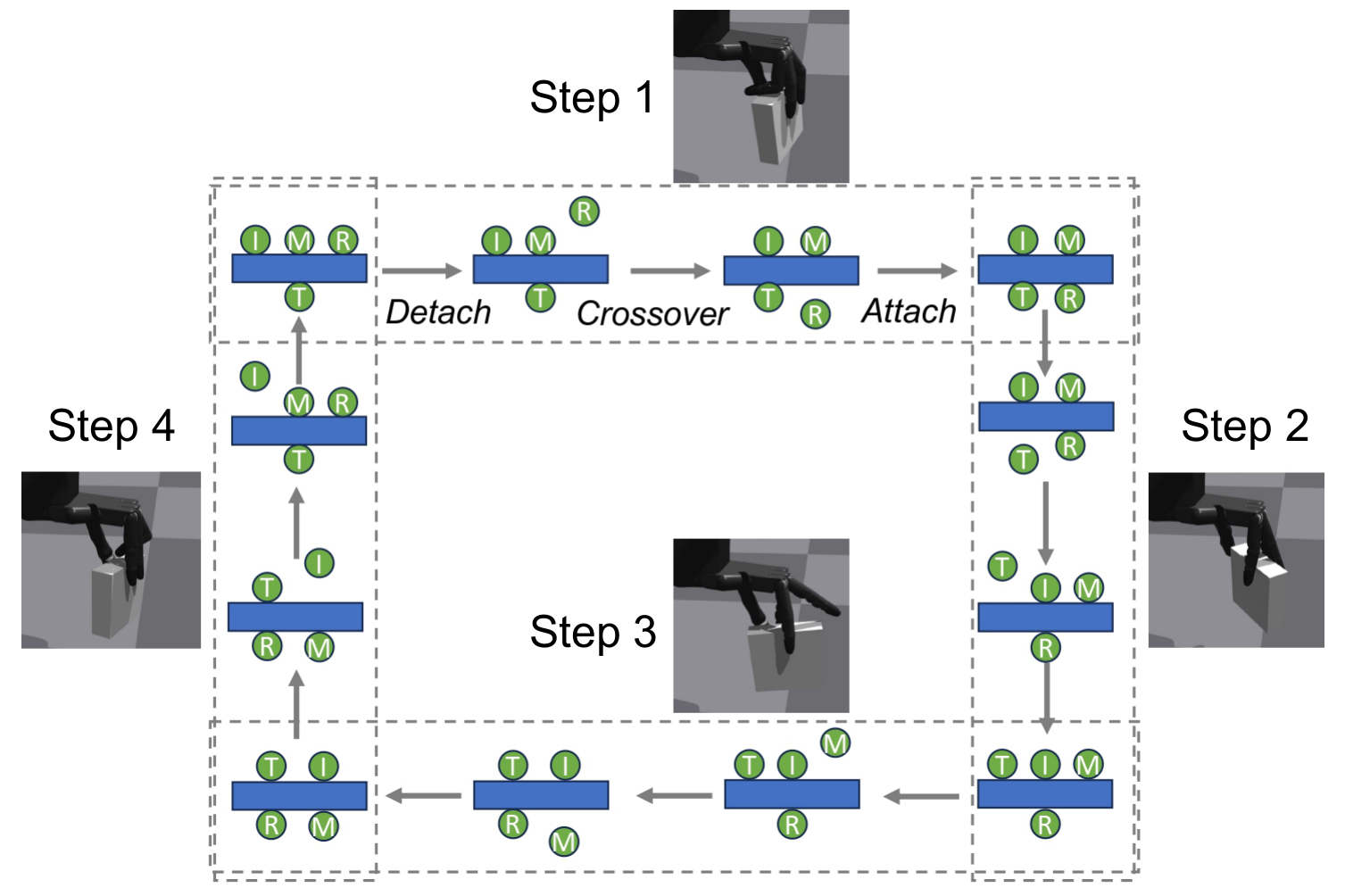}
  \caption{The in-hand rotation skill was decomposed into four sub-skills based on primitive finger motions. Each sub-skill was trained using  RL (image adapted from ~\cite{saito2024apricot}).}
  \label{fig:skilldecomposition}
\end{figure}

The in-hand rotation policies were learned through RL within a IsaacGym Simulator~\cite{makoviychuk2021isaac}. The methodological details including reward functions, are thoroughly described in our previous paper~\cite{saito2024apricot}. In brief, we decomposed the rotation task into four primitive finger motion sub-tasks and trained policies for each sub-task (Fig.~\ref{fig:skilldecomposition}). These policies were designed to eliminate dependence on visual input, relying solely on force feedback, thereby minimizing the sim-to-real gap in the vision domain. Furthermore, we employed a teacher-student learning approach~\cite{qi2023hand} to facilitate training using rich observational information from the simulator. This approach enabled effective policy training for multi-step in-hand manipulation.


\section{Results}
\subsection{Effect of simulation data on the real setup}
First, we evaluated the grasping policy using the Octo model on a real robot (Table~\ref{tab:octo}). The model fine-tuned solely on simulation data failed in real-world experiments, even when the box sizes and shapes used in the real setup were within the range of randomization in the simulation data (see Fig.~\ref{fig:finetuning} for the visual appearances of the tested objects). In contrast, the model that was fine-tuned exclusively on real-world data succeeded with the box used during training (i.e., an orange jello-sized box) but showed inconsistent performance when the box shape or color varied. The model fine-tuned on a combined dataset of simulation and real-world data demonstrated robust results, even when the target objects differed from the ones in the training dataset. These findings suggest that augmenting the dataset with simulation data, incorporating randomized variations in object shape, color, and size, helps to stabilize the model's performance.

\begin{table}[h!]
\caption{Performance of the Octo under different training setups.}
\centering
\begin{tabular}{|c|c|c|c|}
\hline
\textbf{Sim / Real} & \makecell{\textbf{0/40} \\ (real only)} & \makecell{\textbf{4000/40} \\ (proposed)} & \makecell{\textbf{4000/0} \\ (sim only)} \\ \hline
Orange jello-size box  & 5/5  & 5/5  & 0/5 \\ \hline
Green jello-size box   & 0/5  & 5/5  & 0/5 \\ \hline
Blue diamond          & 5/5  & 5/5  & 0/5 \\ \hline
Blue spam-size box    & 1/5  & 5/5  & 0/5 \\ \hline
\end{tabular}
\label{tab:octo}
\end{table}

\subsection{Complete pipeline using a real robot}
\begin{figure}[t]
  \centering
  \includegraphics[width=\columnwidth]{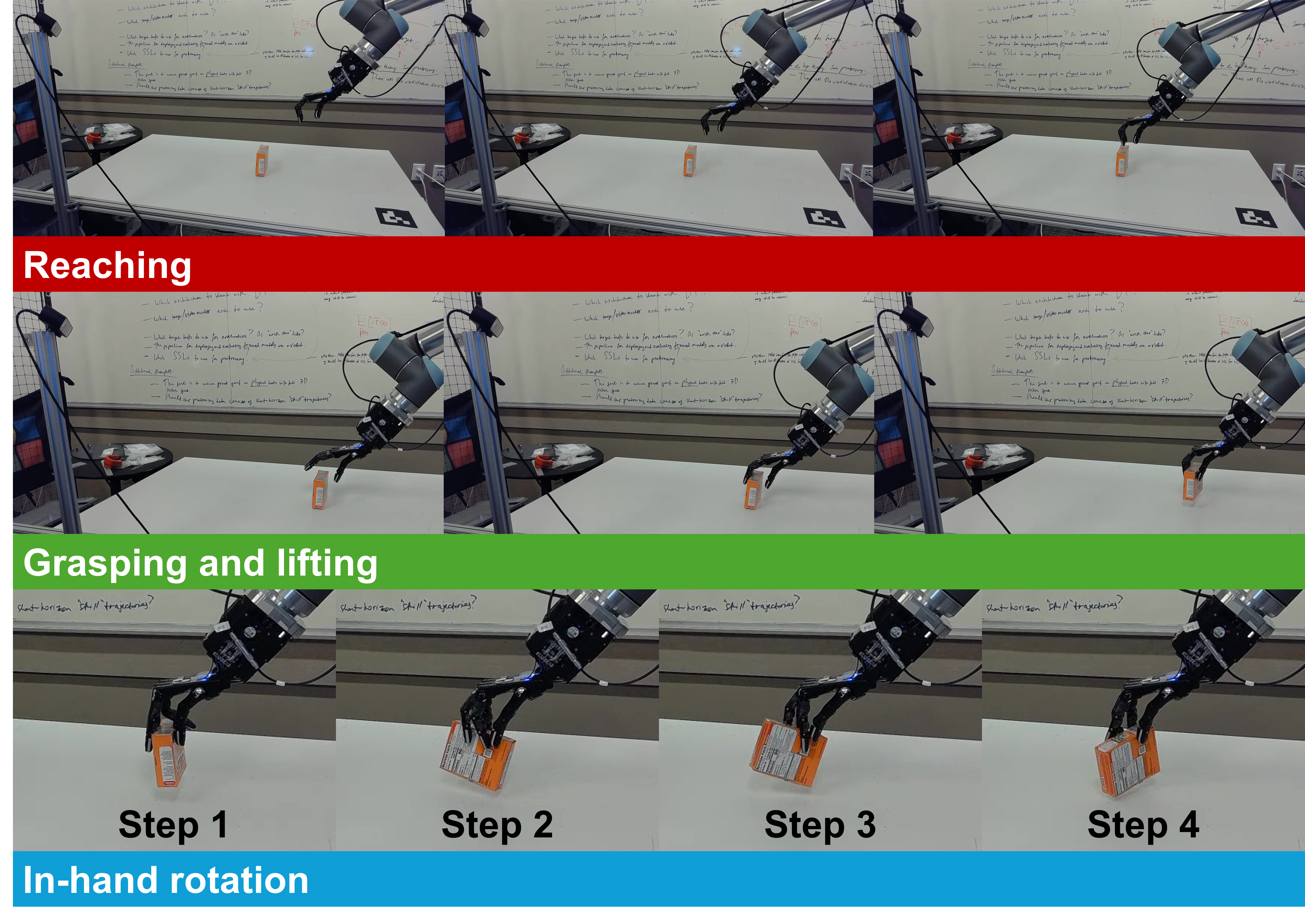}
  \caption{A successful example of end-to-end execution, from reaching to in-hand rotation.}
  \label{fig:end-to-end}
\end{figure}

We tested an integrated pipeline on a real robot system to confirm that our proposed approach is sufficient to solve the target task (Fig.~\ref{fig:end-to-end}). In this example, we provided the text input ``grasp the object from above'' to the Octo model. Once the Octo model determined that the episode had ended, control was transferred from the Octo model to the RL-based skills. The RL skills (Fig.\ref{fig:skilldecomposition}) were executed sequentially, with each skill running for 1000 steps, while commands were sent to the robot at a frequency of 10 Hz.  If the task was deemed a failure through human visual inspection, the process was immediately terminated at that stage. Table\ref{tab:robot_experiment} provides a detailed breakdown of the success rates for the end-to-end experiment. The system successfully advanced through Reaching, Grasp and Lifting, and the first step of in-hand rotation (33/35), and 5/35 trials successfully completed the final rotation step.

\begin{table}[h!]
\caption{Success rate at each step of the multi-step manipulation.}
\centering
\resizebox{\columnwidth}{!}{ 
\begin{tabular}{|c|c|c|c|c|c|c|}
\hline
\textbf{Step} & 
\makecell{\textbf{Reach}} & 
\makecell{\textbf{Grasp} \\ \textbf{\& lift}} & 
\makecell{\textbf{In-hand} \\ \textbf{step 1}} & 
\makecell{\textbf{In-hand} \\ \textbf{step 2}} & 
\makecell{\textbf{In-hand} \\ \textbf{step 3}} & 
\makecell{\textbf{In-hand} \\ \textbf{step 4}} \\ \hline
\textbf{Trials} & 
\makecell{35/35} & 
\makecell{34/35} & 
\makecell{33/35} & 
\makecell{24/35} & 
\makecell{20/35} & 
\makecell{5/35} \\ \hline
\end{tabular}
}
\label{tab:robot_experiment}
\end{table}
\section{Discussion and conclusion}
In this paper, we proposed a modality-driven approach to address multi-step dexterous manipulation. Our method is grounded in neuroscientific insights, utilizing task decomposition based on dominant sensory modalities. The real-world experiments demonstrated the feasibility of our approach with two findings: first, simulation data can substantially enhance the robustness of the VLA model, and second, policies that do not rely on visual information can successfully bridge the sim-to-real gap, enabling reliable execution on physical robots.

There are notable limitations to our approach. First, our method assumes that the learning process is conducted offline, which constrains its applicability in online learning settings. Additionally, even with the simplified problem space achieved by decomposing tasks into sub-skills, the RL policies still rely on tailored reward functions, which limits the scalability of our method. Furthermore, while we demonstrated the feasibility of our approach, the end-to-end performance remains limited. One possible reason is the discrepancy between the grasping state produced by the Octo model after the lifting phase and the distribution assumed during RL training. Optimized domain randomization may mitigate this issue; however, fine-tuning between sub-skills learned using different modalities and methods remains a challenging open question. Despite these constraints, the key contribution of this study lies in its unique perspective on robotic policy design: drawing from neuroscientific principles to decompose complex tasks into sub-skills based on effective modalities.

Our neuroscience-inspired, modality-driven approach provides both a theoretical foundation and a practical pathway for tackling dexterous manipulation tasks, particularly those involving anthropomorphic robotic hands. This work contributes to advancing the field by offering biologically informed insights for addressing the complexities of multi-step robotic manipulation.

\bibliographystyle{unsrt}
\bibliography{bib}

\end{document}